\documentclass[12pt,a4paper]{article}

\usepackage{algorithm}
\usepackage[noend]{algpseudocode}
\usepackage{amsmath}
\usepackage{float}
\usepackage[title]{appendix}
\usepackage{adjustbox,lipsum}
\usepackage{url}
\usepackage{hyperref}
\usepackage{threeparttable}
\usepackage{amssymb}
\usepackage{mathtools}
\usepackage[a4paper,left=2.5cm,right=2.5cm,top=2.5cm,bottom=2.5cm]{geometry}

\makeatletter
\newenvironment{breakablealgorithm}
  {
   \begin{center}
     \refstepcounter{algorithm}
     \hrule height.8pt depth0pt \kern2pt
     \renewcommand{\caption}[2][\relax]{
       {\raggedright\textbf{\ALG@name~\thealgorithm} ##2\par}%
       \ifx\relax##1\relax 
         \addcontentsline{loa}{algorithm}{\protect\numberline{\thealgorithm}##2}%
       \else 
         \addcontentsline{loa}{algorithm}{\protect\numberline{\thealgorithm}##1}%
       \fi
       \kern2pt\hrule\kern2pt
     }
  }{
     \kern2pt\hrule\relax
   \end{center}
  }
\makeatother

\usepackage[inline, shortlabels]{enumitem}
\setlist{itemjoin ={,\enspace},itemjoin* = {,\enspace}}

\begin{document}
\title{Explainable outlier detection through decision tree conditioning}
\author{David Cortes}
\maketitle

\begin{abstract}
This work describes an outlier detection procedure (named \textsc{OutlierTree}) loosely based on the \textsc{GritBot} software developed by RuleQuest research (\cite{gritbot}), which works by evaluating and following supervised decision tree splits on variables, in whose branches 1-d confidence intervals are constructed for the target variable and potential outliers flagged according to these confidence intervals. Under this logic, it's possible to produce human-readable explanations for why a given value of a variable in an observation can be considered as outlier, by considering the decision tree branch conditions along with general distribution statistics among the non-outlier observations that fell into the same branch, which can then be contrasted against the value which lies outside the CI. The supervised splits help to ensure that the generated conditions are not spurious, but rather related to the target variable and having logical breakpoints.
\end{abstract}

\section{Introduction}

This work describes an outlier-detection procedure that aims at producing explanations for why an observation/point can be considered to be anomalous, which are obtained by finding smart conditional distributions of a given variable under which the anomalous observation/point in question would fall according to the conditions, but for which its value on a variable of interest would not match with the distribution of the other observations. These conditional distributions are obtained by splitting/separating/conditioning observations according to some other variable(s) in such a way that the information gain (\cite{c50}) in the variable of interest obtained by splitting the observations (assigning to two or more groups) is maximized, in a similar way as decision tree algorithms such as \textsc{CART} (\cite{cart}) or \textsc{C5.0} (\cite{c50}), which ensure that the conditions that are set for a variable are not spurious, but rather related to the multivariate distribution of the data, and the anomalous value put into context by presenting key information about the variable's distribution among the rest of the observations. An example explainable outlier is sketched below:

\begin{verbatim}
row [2230] - suspicious column: [T3] - suspicious vale: [10.600]
  distribution: 99.951% <= 7.100 - [mean: 1.984] - [sd: 0.750] - [norm. obs: 2050]
  given:
    [query.hyperthyroid] = [FALSE]
\end{verbatim}
(Example was generated from the Hypothyroid dataset\footnote{\url{https://archive.ics.uci.edu/ml/datasets/Thyroid+Disease}}, indicating a case in which a person with very high levels of thyroid hormones was not diagnosed as hyperthyroidal)

The procedure is loosely based on the \textsc{GritBot} software developed by RuleQuest research (\cite{gritbot}), for which to the best of the author's knowledge, no detailed publication about its inner workings has been written, but its source code is published by RuleQuest under a copyleft license and was used as a guide. The implementation of the procedure described here, \textsc{OutlierTree}, is also made open source and freely available\footnote{\url{https://github.com/david-cortes/outliertree}}.

\section{Confidence intervals for outlierness}

If looking only at the distribution of one variable at a time, one logical way of detecting potential anomalous or outlying values is by establishing a possible range of reasonable distribution values through confidence intervals constructed from the mean and standard deviation of the variable in a sample distribution, within which most of the observations should fall with very high certainty, then checking if there is a single or a very small number of observations falling outside of this interval. In this regard, a helpful bound with theoretical guarantees that can be applied to any real-valued distribution is Chebyshev's inequality:

$$
P(|Z_x| \geq k) \leq \frac{1}{k^2} \:\:\:, \:\:\: Z_x = \frac{x - \bar{x}}{\sigma_x}
$$

Which can be used for outlier identification by setting a maximum probability and flagging as outliers any observations which are above or below the mean by the number of standard deviations that result in this probability. This bound is not tight - e.g. in normal distributions the actual probability of getting a value with certain standard deviations above/below the mean will be much smaller than the bound provided by Chebyshev's inequality, which makes an outlier flagged according to this probability even more unexpected.

Some small modifications can help to improve this simple criterion - for example, for long-tailed distributions such as gamma or Poisson, this lower probability bound will be tighter, the standard errors for distribution statistics larger, and as such the threshold after which an observation could be flagged as outlier should ideally be higher for them. In this regard, a possible way of increasing the threshold only for problematic distributions without having to settle for a lower probability is to look at the next smaller/larger observation in sorted order and see if there is a large gap with it, with outliers being flagged as such only if this gap is large and there are not many observations having both a value outside of the range and a large gap with the nearest non-outlier observation. As well, the standard deviations could be calculated with the most extreme values excluded and then adjusted heuristically, in order for it not to be biased by the outliers.

\begin{algorithm}[H]
\caption{FlagOutliers}\label{FlagOutliers}
\hspace*{\algorithmicindent} \textbf{Inputs} $p_{\text{o}}$ (approximate probability of outliers), $Z_{\text{outlier}}$ (minimum z-value for outliers), $Z_{\text{gap}}$ (minimum gap for outlier z-value), $\mathbf{x}$ (sample values)
\begin{algorithmic}[1]
\State Let $n = |\{\mathbf{x}\}|$ (number of obs.)
\State Set number of tail obs. $ n_{\text{tail}} = \lfloor n p_{\text{o}} + 2 n \sqrt{\frac{p_{\text{o}} (1 - p_{\text{o}})}{n}} + 1 \rfloor$
\State Calculate mean and standard deviation excluding the $n_{\text{tail}}$ highest and lowest values $\mu_{\text{adj}}, \sigma_{\text{adj}}$
\State Adjust standard deviation heuristically $\sigma_{\text{adj}} := \frac{n + n_{\text{tail}}}{n - n_{\text{tail}}} \sigma_{\text{adj}}$
\State Sort sample values $\mathbf{x}$
\State Standardize values as $\mathbf{z} = \frac{\mathbf{x} - \mu_{\text{adj}}}{\sigma_{\text{adj}}}$
\For {$i = 1..n_{\text{tail}}$}
	\If {$z_i \leq -Z_{\text{outlier}}$ and $(z_{i+1} - z_i) \geq Z_{\text{gap}}$}
		\State Flag $x_i$ as outlier
	\EndIf
	\If {$z_{n-i+1} \geq Z_{\text{outlier}}$ and $(z_{n-i+1} - z_{n-i}) \geq Z_{\text{gap}}$}
		\State Flag $x_{n-i+1}$ as outlier
	\EndIf
\EndFor
\end{algorithmic}
\end{algorithm}

The default values, taken from \textsc{GritBot}, are as follows: $p_{\text{o}} = 0.01$, $Z_{\text{outlier}} = 8$ (corresponding to $P(|z|) \leq 0.0156$ according to Chebyshev's bound), $Z_{\text{gap}} = 5.33$ (calculated as $Z_{\text{outlier}} - Z_{\text{normal}}$, with $Z_{\text{normal}} = 2.67$).

Transformations on variables whose distributions present large skeweness or long tails can also be applied in order to make distributions closer to normal - for example, taking the logarithm of a power tail distribution results in something that resembles a normal distribution, on which outlier values are easier to determine and the natural gaps between larger-valued observations are smaller. Good candidate transformations could be logarithm for distributions with a large positive skeweness, and exponentiation on distributions with a large negative skeweness (the latter is not used by \textsc{GritBot}, and note that in most cases it will not be able to flag any tail values as outliers). While there is no rule of thumb to determine when does a variable need such type of transformation, and many other potential transformations could also be appropriate, a potential criterion which is aligned with the CI (confidence interval) approach is to always apply a transformation if there is a long tail in the data, as determined by the standard deviation calculated on the central half of the data. The procedure is outlined below:

\begin{algorithm}[H]
\caption{CheckDistTails}\label{CheckDistTails}
\hspace*{\algorithmicindent} \textbf{Inputs} $p_{\text{o}}$ (approximate probability of outliers), $Z_{\text{tail}}$ (maximum expected z-value for extreme observations), $\epsilon$ (minimum value for logarithm), $\mathbf{x}$ (sample values)
\begin{algorithmic}[1]
\State Calculate sample percentiles $x_{p25}$ and $x_{p75}$
\State Calculate central mean and standard deviation $\mu_{\text{cen}}, \sigma_{\text{cen}}$ from $\{ x \in \mathbf{x} \: | \: x_{p25} \leq x \leq x_{p75} \}$
\State Adjust central standard deviation $\sigma_{\text{cen}} := 2.5 \sigma_{\text{cen}}$
\State Let $n = |\{\mathbf{x}\}|$ (number of obs.)
\State Set number of tail obs. $ n_{\text{tail}} = \lfloor n p_{\text{o}} + 2 n \sqrt{\frac{p_{\text{o}} (1 - p_{\text{o}})}{n}} + 1 \rfloor$
\State Sort sample values $\mathbf{x}$
\State Standardize values as $\mathbf{z} = \frac{\mathbf{x} - \mu_{\text{cen}}}{\sigma_{\text{cen}}}$
\If {$z_{n_{\text{tail}}} < -Z_{\text{tail}}$}
	\State Set $\tilde{\mathbf{x}} = \exp( \mathbf{z} )$
	\State Calculate $\tilde{\mu}_{\text{cen}}, \tilde{\sigma}_{\text{cen}}, \tilde{\mathbf{z}}$ from $\tilde{\mathbf{x}}$
	\If {$\tilde{z}_{n_{\text{tail}}} \geq -Z_{\text{tail}}$}
		\State Apply exp-transform to $\mathbf{x}$
	\Else
		\State Flag $\mathbf{x}$ as having a left tail
	\EndIf
\EndIf
\If {$z_{n-n_{\text{tail}}+1} > Z_{\text{tail}}$}
	\State Set $\tilde{\mathbf{x}} = \log( \mathbf{x} - \min{\{ \mathbf{x} \}} + \epsilon )$
	\State Calculate $\tilde{\mu}_{\text{cen}}, \tilde{\sigma}_{\text{cen}}, \tilde{\mathbf{z}}$ from $\tilde{\mathbf{x}}$
	\If {$\tilde{z}_{n-n_{\text{tail}}+1} \leq Z_{\text{tail}}$}
		\State Apply log-transform to $\mathbf{x}$
	\Else
		\State Flag $\mathbf{x}$ as having a right tail
	\EndIf
\EndIf
\end{algorithmic}
\end{algorithm}

(From \textsc{GritBot}, $Z_{\text{tail}} = 5.34$, calculated as $2 Z_{\text{normal}}$, and $\epsilon$ used here was set to $10^{-3}$ for \textsc{OutlierTree})

Note that, while the logic for flagging left/right tails is the same as \textsc{GritBot}, the criteria for whether to apply log-transform or not differs a lot - in \textsc{GritBot} the criteria was as follows, without subtracting the minimum value and without adding a small constant:

\begin{algorithm}[H]
\caption{CheckLogTransform}\label{CheckLogTransform}
\hspace*{\algorithmicindent} \textbf{Inputs} $\mathbf{x}$ (sample values), $\epsilon$ (small positive value)
\begin{algorithmic}[1]
\State Calculate sample percentiles $x_{p25}$, $x_{p50}$, and $x_{p75}$
\If {$\min(\mathbf{x}) > \epsilon$}
	\State Let $R_1 = \frac{\log(x_{p50}) - \log(x_{p25})}{\log(x_{p75}) - \log(x_{p50})}$, $R_2 = \frac{x_{p50} - x_{p25}}{x_{p75} - x_{p50}}$
	\If {$R_2 < 1$ and $|R_1 - 1| < |R_2 - 1|$}
		\State Apply log-transform
	\EndIf
\EndIf
\end{algorithmic}
\end{algorithm}
($\epsilon$ was set to $10^{-6}$)

Just like in \textsc{GritBot}, extreme values belonging to a long tail that did not undergo a transformation are not to be flagged as outliers.

\section{Outliers in categorical data}

The procedure \textsc{FlagOutliers} will however not work for categorical variables in which possible values are unordered discrete categories rather than real numbers. While \textsc{GritBot} will only flag outliers in categorical variables if there is a single dominant category to which most of the observations belong, \textsc{OutlierTree} tries to also flag outliers in situations in which there is no single dominant category, by looking at:
\begin {enumerate*} [(a) ]%
\item the difference in the proportion with respect to the next most-common category \item the prior probability (in the non-conditioned distribution) of the category \item the standard error of the proportion.
\end {enumerate*}

\begin{algorithm}[H]
\caption{FlagOutliersCateg}\label{FlagOutliersCateg}
\hspace*{\algorithmicindent} \textbf{Inputs} $\mathbf{x}$ (sample values), $\mathbf{p}_{\text{prior}}$ (prior probabilities), $n_{\text{prior}}$ (number of observations in the whole data), $Z_{\text{normal}}$ (maximum non-anomalous z-value)
\begin{algorithmic}[1]
\State Let $n = |\{\mathbf{x}\}|$ (number of obs.)
\State Set number of tail obs. $ n_{\text{tail}} = \lfloor n p_{\text{o}} + 2 n \sqrt{\frac{p_{\text{o}} (1 - p_{\text{o}})}{n}} + 1 \rfloor$
\State Let $m = |\{ x \:|\: \exists x \in \mathbf{x} \}|$ (number of categories)
\State Calculate proportions of each category $\mathbf{p}$ from $\mathbf{x}$
\State Set lower bound for expected proportions $\mathbf{p}^{\text{low}} = \min \{ \mathbf{p}_{\text{prior}} - Z_{\text{normal}} \sqrt{\frac{\mathbf{p}_{\text{prior}} (1 - \mathbf{p}_{\text{prior}})}{n_{\text{prior}}}} \:,\: \frac{\mathbf{p}_{\text{prior}}}{2} \}$
\State Sort proportions $\mathbf{p}$
\State Set tail proportion as $m_{\text{tail}} = m$
\For {$i = 1..(m-1)$}
	\If {$(p_{i+1} - p_i) > Z_{\text{normal}} \sqrt{ \frac{ \max \{ p_{i} (1 - p_{i}) \:,\: p_{i+1} (1 - p_{i+1}) \} }{n}}$ and $\frac{p_{i+1}}{2} > p_i$}
		\State Set $m_{\text{tail}} = i$
	\EndIf
\EndFor
\If {$(n \sum_{i=1}^{m_{\text{tail}}} p_i) < n_{\text{tail}}$}
	\For {$i = 1..m_{\text{tail}}$}
		\If {$p_i < p_{i}^{\text{low}}$}
			\State Flag all values $p_i$ as outliers
			\State Terminate procedure
		\EndIf
	\EndFor
\EndIf
\end{algorithmic}
\end{algorithm}

The default value for $Z_{\text{normal}}$ is again $2.67$. The procedure can be extended for new unseen categories by calculating what would happen if a single observation having this new value would be added to the data, with a prior probability given by $\frac{1}{n_{\text{prior}} + 1}$, similar to how it is done by \textsc{GritBot}, in order to use it for flagging outliers outside of the original input data.

One obvious shortfall of \textsc{FlagOutliersCateg} is that it can only flag outliers in sub-sets of the data (otherwise it's impossible for the proportion to fall below the prior probabiltiy). Flagging outliers from categorical data in non-conditional distributions (the whole data) is more challenging, as it's of no use to flag a whole minority category if the proportion is below what was used for Chebyshev's bound (too many false positives), and there is no good answer as for when can a proportion be considered too low. \textsc{OutlierTree} as such uses only these simple rules:
\begin {enumerate*} [(a) ]%
\item can only flag the least common category \item there must be at least 1,000 observations \item there can be at most 1 to 3 observations with the least common value, depending on the sample size \item the next most-common category must have at least 250 observations.
\end {enumerate*}

These criteria differ markedly from \textsc{GritBot}'s condition for flagging categorical outliers in both conditional and non-conditional distributions:
$$
\frac{n - n_{\text{maj}}}{n p_i^{\text{prior}}} \leq \frac{1}{Z_{\text{outlier}}^2}
$$
(With $n_{\text{maj}}$ being the number of obsevations belonging to the majority category)

When comparing these methods, it can be seen that \textsc{GritBot} will only flag outliers when there is a large sample size and a very dominant category to which the majority of the observations belong, whereas \textsc{OutlierTree} will instead flag outliers when there is a proportion which is much smaller than the next one and below what it would be expected in a random sample from the full data, which is more in line with the procedure \textsc{FlagOutliers} for real-valued variables, and is able to flag outliers in smaller sample sizes. As a result, \textsc{GritBot} tends to have a lower recall (too many false negatives) for outliers in categorical variables, while \textsc{OutlierTree} tends to have a lower precision (too many false positives). Note however that \textsc{FlagOutliersCateg} has no theoretical basis for its choices and no correspondence to any probabilistic bound. To the best of the author's knowledge, neither does \textsc{GritBot}'s criterion either.

\section{Conditional distributions}

The \textsc{OutlierTree} procedure aims at finding good conditional distributions of some variable in which to look for outliers and which could be used to explain the logic for flagging an observation as outlier - this is achieved through the same algorithms as decision trees that try to predict a variable of interest (target) based on other variables, applying them to each variable independently after discarding missing values in the target variable.

There has been ample research in induction of decision trees for classification and regression tasks (\cite{cart}, \cite{c50}, \cite{quinlantrees}, among many), aimed at creating predictive models that are explainable or at creating simple models that can be combined to make a better prediction. The general idea is to split a set of points based on the values of one variable being above or below a threshold, in such a way that it makes the obtained groups more homogeneous when measured according to the variable of interest.

Different criteria for how to measure the quality of a split have been proposed, and if the goal is to find good conditional distributions for identifying outliers, it's more logical to choose the type of criterion that favors a more size-balanced split as opposed to some criterion that might result in a very small number of observations being assigned to one branch - e.g. pooled information gain as opposed to un-weighted Gini gain.

As well, since the goal is not to make predictions on the value of the variable on new data, but rather to set transparent rules for conditional distributions, it's still useful to create conditions from splits that would not be usable in typical regression or classification trees, such as assigning observations with missing values to a separate branch instead of typical techniques for handling missing data (\cite{missing}). Categorical variables can also be used more liberally than in the regression/classification scenario - for example, while software such as \textsc{Ranger} (\cite{ranger}) will transform categorical variables to numeric by calculating the mean of the target variable by category at the root node only, in the scenario proposed here these proportions can be recalculated at every tree node instead, resulting in more reasonable splits which could not be used for regression because not every category would be assigned to a branch (if there are no observations of a given category among the sub-sample being split, see \cite{absentlevs}).

Splitting a non-binary categorical variable by another non-binary categorical variable is more complex (see e.g. \cite{multicat}), as it's not possible to calculate a per-category average. Some decision tree algorithms such as \textsc{C5.0} in this case make splits by assigning each category to a separate branch, which is the approach taken by \textsc{GritBot}, but as again the decision trees are not meant to be used for predicting the target variable, it's possible to create different parallel splits based on binarizing the variable multiple times according to whether it is equal to each possible category or not (making it cumulative for ordinal variables), and then using all of these splits independently for distribution conditioning.

While decision trees for classification and regression are based around the idea of choosing the best variable and threshold at each step and then continuing the splitting process recursively, the un-used optimal split thresholds in variables which are not followed for recursive partitioning are still usable for finding outliers in them.

The splitting logic for \textsc{OutlierTree} is sketched as follows:
\begin{breakablealgorithm}
\caption{FindConditions}\label{FindConditions}
\hspace*{\algorithmicindent} \textbf{Inputs} $\mathbf{y}$ (target variable), $\{ \mathbf{x} \}$ (other variables), $g_{\text{min}}$ (minimum gain)
\begin{algorithmic}[1]
\State Initialize $g_{\text{best}} = -\infty$, $v_{\text{best}} = \emptyset$, $c_{\text{best}} = \emptyset$
\If {$\mathbf{y}$ is numeric}
	\For {each other variable $\mathbf{x}$}
		\If {$\mathbf{x}$ is numeric, binary, or ordinal}
			\State Find split point $\text{argmax}_z \sigma^y - \frac{n_l \sigma_l^y + n_r \sigma_r^y + n_u \sigma_u^y}{n} \:\:\text{s.t.}\: l = \{ y_i | x_i \leq z \}, r = \{ y_j | x_j > z \}, u = \{ y_k | x_k \:\text{unknown} \}$ \label{lst:line:numsplit}
		\EndIf
		\If {$\mathbf{x}$ is categorical}
			\State Calculate $\mathbf{p}$ as the mean of $\mathbf{y}$ per category of $\mathbf{x}$
			\State Find split subset $\text{argmax}_q \sigma^y - \frac{n_l \sigma_l^y + n_r \sigma_r^y + n_u \sigma_u^y}{n} \:\:\text{s.t.}\: l = \{ y_i | p_{y_i} \leq q \}, r = \{ y_j | p_{y_j} > q \}, u = \{ y_k | x_k \:\text{unknown} \}$ \label{lst:line:catsplit}
		\EndIf
		\State Calculate gain $g_x = (\sigma^y - \frac{n_l \sigma_l^y + n_r \sigma_r^y + n_u \sigma_u^y}{n}) / \sigma^y$ from optimal split
		\If {$g_x > g_{\text{min}}$}
			\State Find outliers in $\mathbf{y}_l$, $\mathbf{y}_r$, and $\mathbf{y}_u$ separately
			\If {$g_x > g_{\text{best}}$}
				\State Update $g_{\text{best}} = g_x$, $v_{\text{best}} = \mathbf{x}$
			\EndIf
		\EndIf
	\EndFor

\ElsIf {$\mathbf{y}$ is categorical or ordinal}
	\For {each category $c$}
		\If {$\mathbf{y}$ is not ordinal}
			\State Let $\tilde{\mathbf{y}} = \{ \begin{cases}
    			1, & \text{if } y_i = c\\
    			0, & \text{if } y_i \neq c\\
			\end{cases} \forall y_i \in \mathbf{y} \}$
		\Else
			\State Let $\tilde{\mathbf{y}} = \{ \begin{cases}
    			1, & \text{if } y_i \leq c\\
    			0, & \text{if } y_i > c\\
			\end{cases} \forall y_i \in \mathbf{y} \}$
		\EndIf
		\State Calculate base info $I_{\text{base}} = n \log(n) - \sum_{k=0}^{1} |\{ \tilde{y}_i = k\}| \log( |\{ \tilde{y}_i = k\}| )$
		\For {each other variable $\mathbf{x}$}
			\If {$\mathbf{x}$ is numerical or ordinal}
				\State Find split point $\text{argmax}_z I_{\text{base}} - (I_{\text{l}} + I_{\text{r}} + I_{\text{u}})$ from $\tilde{\mathbf{y}}_l, \tilde{\mathbf{y}}_r, \tilde{\mathbf{y}}_u$ as in ~\ref{lst:line:numsplit}
			\EndIf
			\If {$\mathbf{x}$ is categorical}
				\State Calculate $\tilde{\mathbf{p}}$ as the mean of $\tilde{\mathbf{y}}$ per category of $\mathbf{x}$
			\State Find split subset $\text{argmax}_q I_{\text{base}} - (I_{\text{l}} + I_{\text{r}} + I_{\text{u}})$ from $\tilde{\mathbf{p}}_l, \tilde{\mathbf{p}}_r, \tilde{\mathbf{p}}_u$ as in ~\ref{lst:line:catsplit}
			\EndIf
			\State Calculate gain $g_x = (I_{\text{base}}^{\text{orig}} - (I_{\text{l}}^{\text{orig}} + I_{\text{r}}^{\text{orig}} + I_{\text{u}}^{\text{orig}})) / I_{\text{base}}^{\text{orig}}$ from $\mathbf{y}_l, \mathbf{y}_r, \mathbf{y}_u$
			\If {$g_x > g_{\text{min}}$}
				\State Find outliers in $\mathbf{y}_l$, $\mathbf{y}_r$, and $\mathbf{y}_u$ separately
				\If {$g_x > g_{\text{best}}$}
					\State Update $g_{\text{best}} = g_x$, $v_{\text{best}} = \mathbf{x}$, $c_{\text{best}} = c$
				\EndIf
			\EndIf
		\EndFor
	\EndFor
\EndIf
\State Reconstruct $\mathbf{y}_l^{\text{best}}, \mathbf{y}_r^{\text{best}}, \mathbf{y}_u^{\text{best}}$ from $v_{\text{best}}$ and $c_{\text{best}}$, run \textsc{FindConditions} on each
\end{algorithmic}
\end{breakablealgorithm}

The procedure might also add extra conditions that would result in fewer spurious results such as setting minimum sizes for each split, looking for outliers only when a sub-sample has a minimum number of observations, and limiting the decision trees to some pre-determined maximum depth. The default minimum branch sizes in \textsc{OutlierTree} were set as 25 for numerical and 50 for categorical variables, with the minimum sizes required to check for outliers being twice the minimum size for a branch in a split.

The default value for $g_{\text{min}}$ was set as $10^{-3}$ for \textsc{OutlierTree}, which results in fewer viable splits compared to \textsc{GritBot}'s default of $10^{-6}$ which is applied on the raw gain instead (not dividing it the standard deviation nor the information value), thereby compensating for the increased number of flagged outliers brought by other differences between the two.

Applying the gain threshold as a percentage rather than as a raw value like \textsc{GritBot} does has the desirable effect of making the procedure insensitive to the scales of the variables, as well as being able to produce more splits in some variables while avoiding too deep splits in high-variance variables, at the expense of producing sometimes too deep splits in non-informative variables.

\section{Presenting context information}

Once a confidence interval and conditions based on a gain-maximizing split have been determined, the distribution can be described in terms of the mean, standard deviation, value of the largest and smallest non-outlier observations, and percentage of observations (within the sub-sample) which are below/above these values, in order to provide context for a flagged outlier - example:
\begin{verbatim}
row [623] - suspicious column: [age] - suspicious value: [455.000]
  distribution: 99.964% <= 94.000 - [mean: 51.604] - [sd: 18.979] - [norm. obs: 2770]
\end{verbatim}
The high value below which the other observations lie was determined from the \textsc{FlagOutliers} procedure as the z-value with respect to which the outlier observation had a large gap, and the same could be done for outliers that have too small values. Note that the number of observations that are higher/lower than this will always be a minority, provided that $p_o$ is small. Even in case no outliers are found, these statistics for a given set of conditions are still usable when trying to identify outliers in a new sample of data - the largest and smallest values can then be taken as high/low points to contrast against, with thresholds for flagging as outliers set by those values plus/minus $Z_{\text{gap}}$.

In cases of variables that underwent a transformation, the thresholds for flagging outliers can be set from the transformed variable's distribution, while the presented statistics are calculated from the untransformed variable, without incurring any loss of information.

Presentation of the conditions given by a decision tree logic is straightforward, since the branches are defined by them - example from Titanic dataset\footnote{\url{https://www.kaggle.com/c/titanic/data}}:
\begin{verbatim}
row [886] - suspicious column: [Fare] - suspicious value: [29.125]
  distribution: 97.849% <= 15.500 - [mean: 7.887] - [sd: 1.173] - [norm. obs: 91]
  given:
    [Pclass] = [3]
    [SibSp] = [0]
    [Embarked] = [Q]
\end{verbatim}
In this case, \verb_[Pclass] = [3], [SibSp] = [0], [Embarked] = [Q]_ lead to the conditional distribution of "Fare" that was determined by a decision tree of depth 3 (note that some of the variables in  there are ordinal, but when putting the maximum value alone in one branch, the condition can be simplified to "equals"), and again, within each tree node (including the root), the conditions along with high/low thresholds and comparison points for the target variable can be saved to apply them to new data.

While the number of splits that are evaluated by following recursive partitioning is rather large, if only the best splitting variable is followed recursively at each node, the amount of information to remember is manageable for typical datasets with not too many columns (e.g. those from UCI\footnote{\url{https://archive.ics.uci.edu/ml/datasets.php}}), with a space requirement proportional to the square of the number of variables multiplied by the maximum tree depth. While it might also be reasonable to follow each split point at each variable recursively, doing so tends to result in too many outliers for which the conditions oftentimes do not seem logical, and to far larger computation and memory requirements - note for example that in the outlier from the Titanic dataset, the conditions found relate to the quality and location of the accommodation, which is what defines its price, making it a more interesting outlier than if the conditions were random.

\section{Other differences with respect to GritBot}

The \textsc{GritBot} software will add some extra requirements in order to flag an observation as outlier: observations will not be flagged if they differ significantly in some other aspect outside of the criteria used to define a conditional distribution. For example, if the observation has a value in a categorical variable that has an a-priori (non-conditioned) proportion of 25\%, but the proportion is smaller than 2.5\% in the conditional distribution, then the observation will be skipped even if it is outside of some confidence interval and with a gap other than the one in that particular categorical variable. These extra conditions might significantly reduce the number of outliers found in some datasets, which are oftentimes observations that would be desirable to identify. Such conditions were not included in the logic for \textsc{OutlierTree}, which also has the desirable side effect of not making any results dependent on the column order - that is, randomly shuffling the order of the columns will produce the exact same results.

While \textsc{GritBot} will exclude long tails and look for outliers in that variable among observations which do not belong to the tails, \textsc{OutlierTree} will simply forego looking for outliers in the direction in which there is a long tail that is not eliminated by a variable transformation.

When the number of observations is large, \textsc{GritBot} will perform sub-sampling for determining split points, whereas \textsc{OutlierTree}, being developed at a time in which running times for the procedure in commodity hardware are orders of magnitude faster than they were at the time \textsc{GritBot} was created, does not perform any sub-sampling, but rather utilizes more numerically-robust procedures for calculating information gain that allow its correct calculation in large datasets, at the expense of a speed penalty. This also brings the side effect of removing any randomness from the procedure.

Distribution statistics for \textsc{OutlierTree} are calculated with the outliers excluded, and any outliers found in a sub-sample are discarded before splitting that sub-sample recursively, while \textsc{GritBot} will not discard outliers and will calculate distribution statistics with them included.

As well, in cases in which the same observation can be flagged as outlier under multiple tree nodes, \textsc{GritBot} will use the scores (given by Chebyshev's bound and the criterion for categorical outliers) in order to decide which one to prefer, whereas \textsc{OutlierTree} will prefer nodes in this priority order:
\begin {enumerate*} [(a) ]%
\item conditions not involving some value being missing \item number of conditions being smaller \item number of observations in the sub-sample being larger \item outlier scores being more extreme.
\end {enumerate*}

\section{Shortcomings}

There are many aspects that could be handled differently, for example:
\begin{itemize}
	\item Outliers in categorical variables don't tend to stand out when seen as 1-d distributions. Some typical cases of interest for outliers in categorical variables are when one column can only take certain values when another categorical column has another specific value (e.g. "pregnant" and "gender") - these cases might not even be flagged under the procedure proposed here, whereas in other cases it will end up flagging a large number of observations that are not outliers.
	\item The order of categories in ordinal variables is not used when looking for outliers in them, which loses some information.
	\item In multimodal numeric distributions, the same standard deviation is used for both ends, and the difference is counted from the same 1-d mean, which can end up mixing statistics from different underlying distributions and missing some interesting outliers.
\end{itemize}

\section{Conclusions}

This work presented the \textsc{OutlierTree} procedure for explainable outlier detection, inspired by the \textsc{GritBot} software. This procedure is able to provide context in a human-readable format about the outliers that it flags, which are identified as belonging to some logical conditional distribution determined through supervised decision tree splits, and the outlier contrasted against the non-outlier observations in terms of one variable in which it stands out through basic distributional statistics.

While the procedure might not be able to flag all clases of outliers, and its results are perhaps not competitive against those of black-box methods under metrics such as AUC, the outliers found tend to be more helpful for diagnosing problems with the data collection process (e.g. an extra zero in a value) or getting an overview of the most extreme possible combinations of values between different variables.

The methodology is nevertheless able to identify some interesting cases that would be missed by other methods - for example, in the Hypothyroid dataset, a pregnancy at an age of 75 would be flagged as an outlier by \textsc{OutlierTree}, whereas other methods such as local outlier factor (\cite{lof}) or isolation forests (\cite{iso}) would still classify it as normal due to only having a small discrepancy in a binary column (pregnant = 1) whose effect on separability or distance is very small.

\bibliographystyle{plain}
\bibliography{otree}

\end{document}